\pdfoutput=1

\documentclass[11pt]{article}

\usepackage[final]{acl}

\usepackage{times}
\usepackage{latexsym}

\usepackage[T1]{fontenc}

\usepackage[utf8]{inputenc}

\usepackage{microtype}

\usepackage{inconsolata}

\usepackage{graphicx}

\usepackage[nolist]{acronym}
\usepackage{multirow}

\begin{acronym}
    \acro{rq}[RQ]{research question}
    \acroplural{rq}[RQs]{research questions}
    \acro{nlp}[NLP]{natural language processing}
    \acro{llm}[LLM]{large language model}
    \acro{lora}[LoRA]{low-rank adaptation}
    \acroplural{llm}[LLMs]{large language models}
    \acro{ai}[AI]{artificial intelligence}
    \acro{qa}[QA]{question answering}
    \acro{kbqa}[KBQA]{question answering over knowledge bases}
    \acro{nlu}[NLU]{natural language understanding}
    \acro{nlg}[NLG]{natural language generation}
    \acro{kg}[KG]{knowledge graph}
    \acroplural{kg}[KGs]{knowledge graphs}
\end{acronym}

\title{Conversational Exploratory Search of Scholarly Publications \\ Using Knowledge Graphs}

\author{Phillip Schneider \and Florian Matthes \\
         Technical University of Munich, Department of Computer Science, Germany \\
         \texttt{\{phillip.schneider, matthes\}@tum.de}}

\begin{document}
\maketitle
\begin{abstract}
Traditional search methods primarily depend on string matches, while semantic search targets concept-based matches by recognizing underlying intents and contextual meanings of search terms. Semantic search is particularly beneficial for discovering scholarly publications where differences in vocabulary between users' search terms and document content are common, often yielding irrelevant search results. Many scholarly search engines have adopted knowledge graphs to represent semantic relations between authors, publications, and research concepts. However, users may face challenges when navigating these graphical search interfaces due to the complexity and volume of data, which impedes their ability to discover publications effectively. To address this problem, we developed a conversational search system for exploring scholarly publications using a knowledge graph. We outline the methodical approach for designing and implementing the proposed system, detailing its architecture and functional components. To assess the system's effectiveness, we employed various performance metrics and conducted a human evaluation with 40 participants, demonstrating how the conversational interface compares against a graphical interface with traditional text search. The findings from our evaluation provide practical insights for advancing the design of conversational search systems.
\end{abstract}

\section{Introduction}
\label{sec:introduction}

Digital publication platforms have greatly expanded the accessibility of scholarly articles, offering an extensive range of publications that can be efficiently discovered through integrated search engines. These digital platforms provide researchers with access to millions of scholarly documents, encompassing conference papers, journal articles, workshop proceedings, and book chapters. The number of scholarly documents is growing exponentially, with estimates suggesting that it doubles approximately every 15 years \cite{Bornmann2021GrowthRO}. As the body of literature grows, traditional keyword-based search methods are becoming less effective at filtering and ranking relevant documents. These lexical methods rely heavily on well-formulated queries and otherwise yield irrelevant results. Researchers are often hindered by the so-called \textit{vocabulary mismatch problem}, which manifests as differences between search terms and the terminology in the documents \cite{furnas1987vocabulary}. This issue is especially pronounced in open-ended and exploratory search scenarios, where users navigate unfamiliar information spaces. In such scenarios, users' incomplete knowledge of certain topics prevents them from formulating queries to access the information they need \cite{schneider2023investigating}.

Reacting to the challenges posed by the high volume of scientific output, digital publication platforms have begun to make their search functionalities more intelligent by employing semantic search methods using \ac{nlp}. These methods enable search engines to understand the context and intent behind user queries. Moving beyond exact keyword matches to semantic matches on a conceptual level can help identify relevant articles, even when different terms are used, thereby aiding users in discovering publications from subfields they are unfamiliar with. Complementing this, \acp{kg} have established themselves as a versatile data structure for representing semantic relationships between interconnected entities like institutions, authors, topics, research fields, and other concepts.

Two popular examples of platforms that have incorporated \acp{kg} are Microsoft Academic \cite{wang2020microsoft} and Semantic Scholar \cite{kinney2023semantic}. Microsoft Academic created the Microsoft Academic Graph, which supports semantic search, contextual query understanding, and personal recommendations. Similarly, the Semantic Scholar platform operates on the Semantic Scholar Academic Graph with more than 200 million papers. While these platforms offer a range of graph-based features and visualizations, they introduce usability hurdles by rendering graphical search interfaces more complex. Graphical interfaces can become less effective for exploratory search because of the added layers of complexity, causing users to experience cognitive overload \cite{sweller1988cognitive}. This might be exemplified by the decline and eventual termination of Microsoft Academic in 2021, whose intricate interface likely has contributed to deterring users \cite{orduna2014silent}.

To address the complexity of graphical semantic search interfaces, we propose developing a conversational interface for discovering scholarly publications via dialogue interactions, leveraging a \ac{kg} data structure. The emerging paradigm of conversational search promises to satisfy information needs using intuitive information-providing conversations while avoiding information overload \cite{radlinski2017theoretical}. Through interactions with conversational agents, users can resolve ambiguities, refine their queries, narrow down the relevant search space, and extract novel insights. Our study aims to provide insights into how conversational search systems integrated with \acp{kg} can enhance the discovery of publications, thereby improving navigation and information retrieval in the scholarly research landscape. To demonstrate the effectiveness of our developed system, we utilize the open-source corpus of the ACL Anthology as our data foundation. The source code, models, datasets, and questionnaires are made available via a public GitHub repository.\footnote{\href{https://github.com/philotron/CS-Scholarly-KG}{Repository: github.com/philotron/CS-Scholarly-KG}} Our three main contributions are as follows: (1) We propose an architecture for integrating a conversational exploratory search system with a scholarly \ac{kg}. (2) We implement the system by assembling different task-specific language models. (3) We conduct both a model-centric performance assessment and a human evaluation of the developed system with 40 participants.

\section{Related Work}
\label{sec:rel-work}
Conversational search systems are defined as conversational interfaces that support acquiring information through multi-turn dialogues. These systems progressed significantly in recent years, largely driven by the rapid adoption of \acp{llm}. A growing body of research focuses on augmenting conversational search systems with \acp{llm} \cite{schneider-etal-2024-engineering}, including utterance understanding \cite{kuhn2023clam}, dialogue management \cite{friedman2023leveraging}, knowledge retrieval \cite{lewis2020retrieval}, and response generation \cite{sekulic2024towards,schneider-etal-2024-comparative}. While \acp{llm} hold great potential for conversational search systems, they are not without shortcomings. \acp{llm} can hallucinate or omit crucial information, and their outputs often lack transparency regarding the source of generated content \cite{ji2023survey}. In addition, \acp{llm} are usually non-deterministic, posing challenges in ensuring consistent and correct knowledge due to the randomness in their text generation processes. 

\begin{figure*}[h]
\centering
  \includegraphics[width=\textwidth]{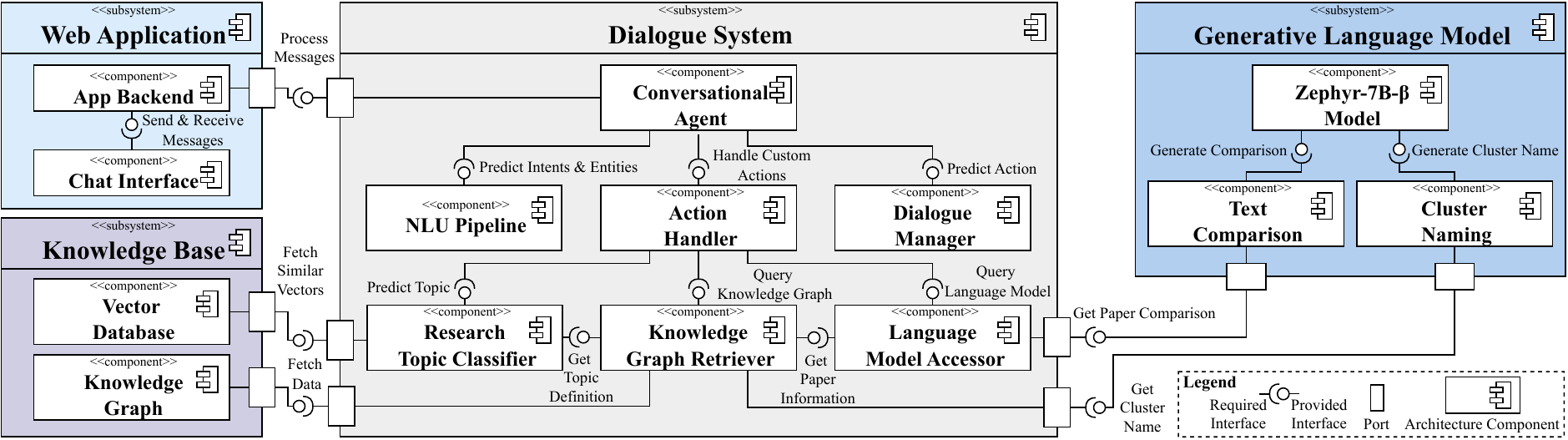}
  \caption{{Architectural components of the conversational exploratory search system.}}
  \label{fig:architecture}
\end{figure*}

To mitigate issues of factuality and reliability in conversational systems and \acp{llm}, researchers have studied using \acp{kg} to ground outputs in verifiable data sources. Integrating \acp{kg} with dialogue systems has long been a focus in the literature. \acp{kg} can replace static domain knowledge with dynamic ontologies and have shown effectiveness in conversational \ac{qa} \cite{christmann2019look,schneider2024evaluating}. By navigating entity nodes and relationships, \acp{kg} enhance conversational context and information exploration. Numerous studies support the use of \acp{kg} for improving utterance understanding, response generation, and dialogue management \cite{chen2019kbrd,chen2023graph}. While \acp{kg} are increasingly being combined with conversational agents in fields such as healthcare, law, and business, there remains a significant gap in their application within the scholarly domain. Thus far, \citet{meloni2023integrating} are the only ones to propose combining a conversational agent with the \textit{Academia/Industry DynAmics} (AIDA) \ac{kg} \cite{angioni2021aida}. Their AIDA chatbot focuses on QA by executing database queries to count, list, compare, or describe scholarly entities (e.g., authors or conferences), thereby offering senior researchers an overview of the research landscape through bibliometric data \cite{meloni2023aida}. 

In contrast, our proposed system supports the conversational discovery of research articles for users with vague goals in open-ended search scenarios, building on insights from our previous work \cite{schneider-etal-2023-data}. Therefore, unlike the AIDA chatbot, which primarily assists senior researchers, our proposed system is designed to support exploratory information search for non-expert users looking to discover relevant publications on a given topic without requiring in-depth knowledge. 

\section{Conversational Search System}
\label{sec:sys}

\subsection{System Architecture}
\label{sec:sys-architecture}
The developed dialogue system helps users narrow down relevant publications via a three-phase search process. An overview of the conversation flow is illustrated in Figure~\ref{fig:dialogue-states} in Appendix~\ref{sec:appendix}. In the first phase, the system receives a short description of a search goal (e.g., ``I want to study how people express their feelings on social media.'') and then it assists users by recommending an appropriate \ac{nlp} research topic to explore (e.g., \textit{Emotion Analysis}). Second, users can iteratively choose thematic clusters of articles within the selected research topic (e.g., \textit{Emotion Detection in Social Media Text}). Third, users are presented with a list of articles at the lowest cluster level, where they can compare papers based on summarized information from the abstracts. Finally, users have the option to either access links to the full texts of the papers or continue exploring other research topics, clusters, or articles. In Section~\ref{sec:model-results}, we will elaborate on the developed \ac{nlp} models powering the three-phase search process through (1) topic classification, (2) text clustering, as well as (3) text summarization.

Figure~\ref{fig:architecture} illustrates the system architecture, which is structured into four distinct subsystems. Each subsystem encompasses multiple components responsible for typical dialogue system functions. The front end is a conversational interface implemented as a web application subsystem using the open-source framework Streamlit.\footnote{\href{https://streamlit.io/}{Streamlit: https://streamlit.io}} It features a basic chat interface with a text input form and a scrollbar for the current dialogue history. User messages entered in the chat interface are sent to a dialogue system built with RASA, an open-source machine learning framework.\footnote{\href{https://rasa.com/}{RASA: https://rasa.com}} RASA supports the development of conversational agents with a \ac{nlu} pipeline that recognizes intents and entities from user utterances. Based on these semantically parsed user utterances, the agent's dialogue manager, which takes into account dialogue states, dialogue policies, and conversation context, predicts the system's next actions. Aside from standard actions like producing a simple response, the agent connects with an action handler component to implement custom actions. One custom action is the \ac{kg} retriever component. It enables the construction and execution of structured queries to retrieve data from the scholarly \ac{kg}, such as abstracts, thematic clusters, or research topics. It connects with the knowledge base subsystem, which hosts the \ac{kg} in a Neo4j property graph database.\footnote{\href{https://neo4j.com/}{Neo4j: https://neo4j.com}} For the query construction, extracted entities from user utterances are matched with those existing in the \ac{kg} to fill out template queries. 

In addition to the \ac{kg}, the knowledge base subsystem hosts the open-source vector database Weaviate, which performs embedding-based similarity search.\footnote{\href{https://weaviate.io/}{Weaviate: https://weaviate.io}} Together with the \ac{kg}, the vector database supports the research topic classifier component by finding the closest research topics from user inquiries. Another component that powers a custom action is the language model accessor, which provides a connecting endpoint to the generative language model subsystem. Inside this subsystem, we host the open-source \ac{llm} Zephyr-7B-Beta \cite{tunstall2023zephyr}. The subsystem offers two inference endpoints for dynamic prompting. The endpoints are used to generate names for paper clusters and summarized paper comparisons.

The described system is deployed on three virtual machines (VMs) in a cloud environment. The first VM operates the dialogue system that interacts directly with users through the conversational interface. The second VM acts as a database server, while the third VM, equipped with a GPU (16 GB memory), hosts the large language model. Despite the architecture's various technical components depicted in Figure~\ref{fig:architecture}, the conversational interface hides the complexity of the underlying \ac{kg}, providing a highly accessible search experience.

\subsection{Knowledge Graph and Vector Database}
\label{sec:sys-kg}

To establish the data foundation for the conversational search system, we constructed a domain-specific \ac{kg} with over 85,000 research articles sourced from the ACL Anthology.\footnote{\href{https://aclanthology.org/}{ACL Anthology: https://aclanthology.org}} A compact overview of the data schema with entity nodes and relations is presented in Figure~\ref{fig:data-model} in Appendix~\ref{sec:appendix}.

Sourcing articles from the ACL Anthology provided us with detailed metadata on authors, venues, and publication years that were automatically transformed into nodes and relations of the \ac{kg}. In addition, we assigned each article to one or multiple research topics. To achieve this, we used a previously established taxonomy of \ac{nlp} research topics from \citet{schopf-etal-2023-exploring}, which is organized as a two-level hierarchy with main topics and subtopics, along with topic names and human-written definitions. This taxonomy includes 12 main topics, such as \textit{Text Generation} or \textit{Sentiment Analysis}, and a total of 71 subtopics (e.g., \textit{Question Generation} or \textit{Emotion Analysis}). For classifying articles, we employed two fine-tuned language models for classifying publications: a SPECTER2-based model \cite{singh-etal-2023-scirepeval} for multi-label topic classification based on the used \ac{nlp} taxonomy \cite{schopf-etal-2023-exploring,schopf2024nlpkg}, and a SciNCL-based model \cite{ostendorff-etal-2022-neighborhood} to classify if a publication is a survey paper consolidating information from several other publications. The taxonomy inside the \ac{kg} is later applied to train a classification model that predicts a relevant \ac{nlp} subtopic based on a described search goal. By providing topic definitions and listing related topics, the conversational agent can guide users through the \ac{nlp} taxonomy.

While effective for broad classification, the two-level \ac{nlp} taxonomy is not granular enough to account for thematic differences within a given subtopic. For example, the subtopic  \textit{Emotion Analysis} includes over 780 publications, which share only a few common characteristics. To have a more fine-grained filtering mechanism, we clustered papers based on their title and abstract content (e.g., specific techniques, application domains, or benchmark datasets). These thematic clusters are pre-computed and modeled as nodes in the \ac{kg}. The clustering and cluster naming methods will be discussed in more detail in Section~\ref{sec:model-results}.

In addition to the \ac{kg} database, we installed a vector database that supports various embedding models and similarity metrics, making it ideal for efficiently ranking semantically similar documents. We employed the SPECTER2 embedding model \cite{singh-etal-2023-scirepeval} for generating vectors from papers' titles and abstracts and used them for the mentioned research topic classification, mapping \ac{nlp} topics from the taxonomy to user requests during the search dialogue in real-time. A document identifier in the vector database links these embeddings to the papers in the \ac{kg}. As a last construction step, we further enriched the \ac{kg} with metadata from the Semantic Scholar API, including one-sentence \textit{too long; didn't read} (TLDR) summaries, citation counts, and publication references.

\section{Results and Discussion}
\label{sec:results}

\subsection{Model Training and Evaluation}
\label{sec:model-results}

\paragraph{Research Topic Classification.}
In the following sections, we report the results of training and evaluating the \ac{nlp} models that underpin the three-phase search process of our developed dialogue system: (1) research topic classification, (2) article text clustering, and (3) comparative article summarization.
The first phase involves classifying an uttered search goal or problem description into a fitting \ac{nlp} research topic. This is especially helpful for users in exploratory search settings because they may not be familiar with all existing fields of study and struggle to phrase their queries using the correct terminology. Due to the absence of datasets that map search goals expressed in layman's terms to \ac{nlp} topics, we created a synthetic multi-class dataset using GPT-3.5-Turbo (version: 0613) \cite{openai2022chat}. We prompted the \ac{llm} to generate questions on the 12 main topics in our taxonomy using three distinct personas: a computer science student with only peripheral \ac{nlp} knowledge, a businessperson with practical experience of \ac{nlp} tools but minimal technical expertise, and a non-technical, non-academic individual whose technology use is limited to basic tasks. Persona-specific prompting yielded diverse inquiries in layman's language. For example, the question ``How are computers able to respond when we ask them questions?'' was generated for the topic \textit{Natural Language Interfaces}. To account for questions unrelated to \ac{nlp}, we also generated a set of out-of-scope questions such as ``Who discovered the laws of thermodynamics?'' Following a quality inspection of the synthetically produced questions, we assembled a training dataset of 1601 examples, consisting of 120 questions for each of the 12 topics and 161 general questions. We also derived a test dataset containing 364 examples with a balanced class distribution similar to the training dataset.

In our experiments, we evaluated three classification approaches: vector similarity search, prompting a \ac{llm} (GPT-3.5-Turbo), and few-shot fine-tuning of a transformer model with the SetFit framework \cite{Tunstall2022EfficientFL}. Concerning the vector search approach with the SPECTER2 model, we measured the cosine similarity to compare vectors of embedded user queries with paper embeddings in our vector database to retrieve the 100 most similar papers. We found that a similarity threshold below 77\% effectively filters out the non-\ac{nlp}-related questions. Using the scholarly \ac{kg}, we aggregated linked topics for these papers and predicted the most frequent topic as output class. For the \ac{llm} approach, we crafted a zero-shot prompt for GPT-3.5-Turbo, provided in Appendix~\ref{sec:appendix}, which instructed the \ac{llm} to classify the appropriate topic from the list of 12 main topics or answer with ``None'' if the question was not related to \ac{nlp}. Moreover, we tested the SetFit approach for fine-tuning the sentence transformer model multi-qa-MiniLM-L6-cos-v1 \cite{wang2020minilm}. We trained for 3 epochs, a batch size of 16, and 30 SetFit iterations for contrastive learning. 

\begin{figure}[h]
\centering
  \includegraphics[width=\linewidth]{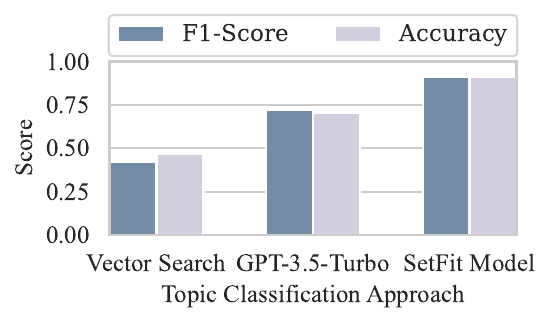}
  \caption{Comparison of accuracy and F1-scores for three topic classification approaches.}
  \label{fig:topic-results}
\end{figure}

Figure~\ref{fig:topic-results} illustrates the classification performance for each approach. While vector search achieved a macro F1-score below 0.50., GPT-3.5-Turbo achieved a score near 0.75; however, it exhibited a bias toward particular topics, leading to overprediction and incorrectly classifying general questions as \ac{nlp} topics. The SetFit model demonstrated superior performance over the two other approaches with a score of 0.95. Consequently, we implemented the topic classifier component with this fine-tuned model for main topic classification in combination with similarity search for classifying the subtopic. This allows a more nuanced classification of user queries into subtopics, given the more detailed information in the paper abstracts.

\paragraph{Article Text Clustering.}
After selecting an \ac{nlp} subtopic, the conversational search agent guides users by presenting clusters of papers to further narrow down the search space. We tried out various clustering methods for thematically grouping similar papers because listing all papers within one subtopic at once is impractical. Selecting thematic clusters makes it easier for users to find relevant papers by iteratively choosing smaller, more specific clusters. Our experiments indicated that agglomerative clustering, a hierarchical bottom-up clustering approach, was the most effective in producing mutually exclusive clusters at each hierarchy level \cite{murtagh2012algorithms}. We employed the SPECTER2 embeddings of the publications, which were the same ones used for the similarity search as part of the research topic classification.

Initially, we used a distance threshold of 10, resulting in clusters averaging 15 publications each. However, this led to too many clusters inside a given research topic, making cluster selection very cumbersome. The distance threshold represents the maximum distance within which items are clustered together. To improve the clustering, we adopted an iterative hierarchical approach. We progressively decreased the distance threshold at each cluster level, keeping the number of clusters small while increasing the paper similarity within each subcluster to facilitate user navigation. Clustering stopped when fewer than 10 publications remained per cluster, ensuring a user-friendly number to display. Overall, we constructed a granular hierarchy of 47,035 thematic clusters, which were modeled as nodes in the constructed scholarly \ac{kg}.

Next, it was necessary to assign human-readable cluster names to help users identify relevant clusters during the conversational search interaction. We applied a term frequency-inverse document frequency (TFIDF) vectorizer to extract important words from the titles of all publications within each cluster. Through several experiments, we found that an n-gram range of (2,5), considering sequences of two to five consecutive words, yielded good cluster names. However, we observed that a few clusters had identical names. To resolve this, we performed another cluster naming iteration, taking into account all previous names. If a name was repeating, we selected the second or third most relevant TFIDF label to ensure unique names. While the TFIDF-derived names were readable, they often contained too much detail and domain-specific words, rendering them less accessible to non-expert users. To make cluster names more understandable, the aforementioned Zephyr-7B-Beta \ac{llm} was applied. Within the \ac{llm} subsystem in our architecture, a specialized component was developed for cluster naming. A dynamic prompt, detailed in the GitHub repository, was created to transform an existing TFIDF cluster name alongside five randomly selected paper titles from a chosen cluster into a more comprehensible cluster name (e.g., \textit{Emotion Detection in Social Media Text} or \textit{Extraction of Concept Maps for Multi-Document Summarization}). To minimize response latency during a conversation, names for clusters and subclusters were pre-computed and stored in the scholarly \ac{kg}.  

\paragraph{Comparative Text Summarization.}
In the last phase of the conversational search process, users can compare papers listed at the lowest cluster level. Although these papers are already thematically related, the comparison allows users to discern specific similarities and differences, aiding in determining which paper to read more thoroughly. The language model subsystem allows for the summarization of objectives and results of two selected papers, which are generated in real-time upon request with Zephyr-7B-Beta. Given that injecting full abstracts can impede the \ac{llm}'s ability to accurately detect objectives and results, only relevant portions of the abstract are provided in a dynamic prompt, which has been shown to reduce hallucinated outputs \cite{martino2023knowledge}. 

To this end, we first classified abstract sentences that discuss objectives or results using {SciBERT}, a language model pre-trained on scientific text \cite{beltagy-etal-2019-scibert}. We fine-tuned {SciBERT} on a labeled dataset from \citet{gonccalves2020deep}, including 500 computer science abstracts and 3,287 sentences classified as background, methods, objectives, results, or conclusions. After hyperparameter optimization, our fine-tuned model achieved an F1-score of 75.39\%, which is around one percentage point higher than the model from \citet{gonccalves2020deep} with 74.60\%. Finally, we applied our model to all the publication abstracts in our \ac{kg} and stored the classified objectives and results sentences accordingly. More details about our fine-tuned model are available in the repository. A dynamic \ac{llm} prompt for text summarization was crafted, as shown in Table~\ref{tab:prompts} in Appendix~\ref{sec:appendix}. Two researchers manually assessed the generated comparisons. Initial experiments with other models, such as Falcon-7B and Llama-2-7B, showed that these models were less attuned to following the instruction, often producing hallucinated content or excessively verbose responses, making them unsuitable for conversational interactions. As a result, we selected Zephyr-7B-Beta, which delivered better output in terms of style and faithful content.

\begin{table*}[t]
\small
\centering
\begin{tabular}{l|c|c|c|c}
\hline
\multirow[t]{2}{*}{\textbf{Evaluation Metric}} & \multicolumn{2}{c|}{\textbf{Scenario 1}} & \multicolumn{2}{c}{\textbf{Scenario 2}} \\ 
\cline{2-5}
\textbf{Mean (Std. Dev.)} & \textbf{Conversational} & \textbf{Graphical} & \textbf{Conversational} & \textbf{Graphical} \\
\hline
System usability scale & 76.00 (18.94) & 77.25 (15.28) & 76.63 (16.63) & 65.25 (23.91) \\ 
\hline
Readability & 4.50 (0.95) & 3.40 (1.14) & 4.45 (0.76) & 3.20 (1.54) \\
\hline
Correctness & 4.25 (0.97) & 4.05 (1.00) & 4.25 (0.72) & 3.85 (1.31) \\
\hline
Usefulness & 4.50 (0.61) & 3.65 (0.99) & 4.30 (0.80) & 2.95 (1.23) \\
\hline
Summary quality & 4.10 (0.85) & - & 4.15 (0.67) & - \\
\hline
Overall satisfaction & 4.15 (0.88) & 3.45 (1.00) & 4.10 (1.07) & 2.85 (1.14) \\
\hline
\end{tabular}
\caption{Overview of mean and standard deviation of evaluation metrics by search scenario and interface type.}
\label{tab:human-evaluation}
\end{table*}

\subsection{Human Evaluation}
\label{sec:eval-results}
\paragraph{Experiment Design.}
To evaluate the three-phase search system in an end-to-end manner, we designed a user study in which participants explored publications related to two predefined search scenarios. They interacted with the conversational search interface and a graphical interface featuring a traditional text-based search, allowing us to compare the effectiveness of both systems. For the experiment, we recruited 40 participants from university courses and social networks according to criteria that match our target user group of non-experts. Table~\ref{tab:demographics} in Appendix~\ref{sec:appendix} gives an overview of participant demographics. All participants had at least basic technical knowledge, good English proficiency, and an interest in \ac{nlp} without having expert-level knowledge. The gender composition was 35\% female and 65\% male, ranging in age from 20 to 29 years, with an average age of 25. 

Prior to the user experiment, we randomly assigned participants into two groups (Group A and Group B), which determined the sequence in which each group used the search interfaces for the two search scenarios (Scenario 1 and Scenario 2). We ensured that the demographic characteristics of both groups were similarly distributed. Group A is exposed to Scenario 1 with the conversational interface first, followed by Scenario 2 with the graphical interface. Conversely, Group B is exposed to Scenario 1 with the graphical interface first, then Scenario 2 with the conversational interface. This crossover design allows each participant to test both interfaces but for a different scenario to avoid learning effects. Scenario 1 is about analyzing emotional expressions on social media related to mental health during the COVID-19 pandemic, while Scenario 2 focuses on creating multiple-choice exams for a programming course. Both scenarios end with the instruction: ``Your task is to use the provided search interface below to find papers related to the described scenario.'' The full scenario descriptions are documented in Table~\ref{tab:scenario} in Appendix~\ref{sec:appendix}.

\begin{figure*}[h]
\centering
  \includegraphics[width=\textwidth]{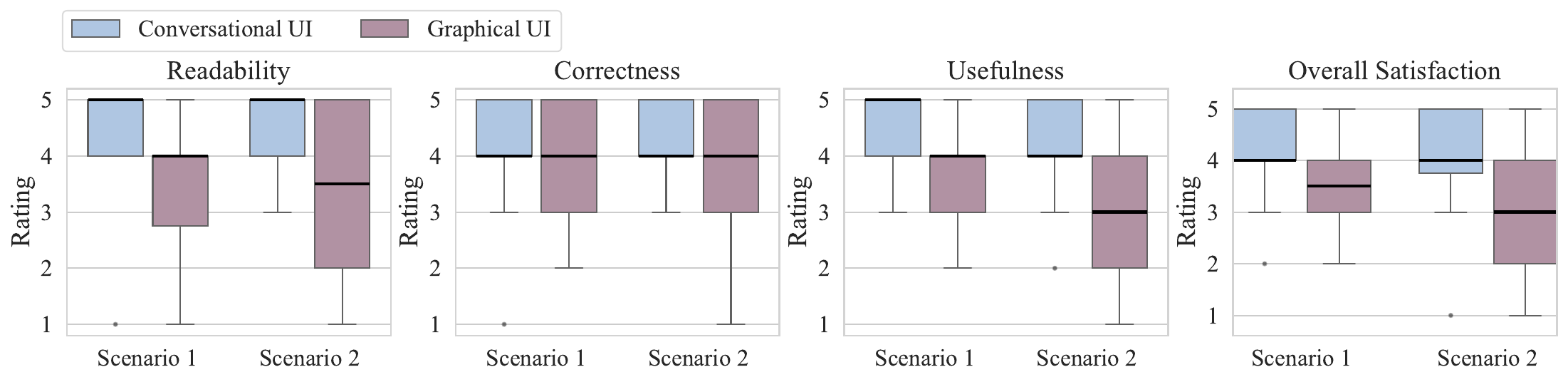}
  \caption{Comparison of rating distribution between the conversational and graphical user interface (UI) for readability, correctness, usefulness, and overall satisfaction. The thick line intersecting the box marks the median.}
  \label{fig:box-plot}
\end{figure*}

Participants were given approximately 10 minutes to interact with the interfaces (see screenshots in Table~\ref{tab:screenshots}), followed by an evaluation questionnaire, which we share in the repository. The latter includes 10 questions from the \textit{system usability scale} (SUS) \cite{brooke1996sus}. The SUS metric is calculated using a specific formula, resulting in a value between 0 and 100, with 68 being considered average. Furthermore, participants were asked five questions to rate the general information quality in terms of readability, correctness, and usefulness, the quality of the generated comparisons, as well as overall satisfaction. The questions were answered on a 5-point Likert scale, where a rating of 5 denoted the most favorable value. In addition, we included two open-ended free-text fields for feedback on the system's strengths and weaknesses.

\paragraph{Quantitative Analysis of Evaluation Metrics.}
Based on the questionnaire responses, we conducted both quantitative and qualitative analyses. Table~\ref{tab:human-evaluation} lists the mean and standard deviation for each evaluation metric grouped by scenario and interface. We found that all data points were within reasonable ranges without containing significant outliers. Generally, the ratings for the conversational interface tend to be more favorable across the various evaluation metrics. This is also reflected in the overall satisfaction scores for Scenario 1 and Scenario 2, with ratings of 4.15 and 4.10 for the conversational interface compared to 3.45 and 2.85 for the graphical interface. The conversational interface especially surpasses the graphical interface in readability and usefulness metrics, as reflected by mean ratings that were around one point higher across both scenarios. We hypothesize that the dialogue interaction was not as overwhelming, delivering information in a more digestible format and increasing its overall utility by offering additional choices for paper selection. This positive feedback was likely also influenced by the quality of summarized paper comparisons, which achieved solid scores of 4.10 in Scenario 1 and 4.15 in Scenario 2.

Inspecting the perceived system usability, the SUS scores for Scenario 1 were similar for both interfaces, with SUS scores at 76.00 and 77.25. Since all participants presumably had prior experience on how to use a text-based search engine as well as a standard chat interface, it is not surprising that the scores are similar. It is very likely that each participant was already acquainted with operating both types of interfaces. In Scenario 2, the conversational interface maintained a comparable score of 76.63, whereas the graphical interface had a lower score of 65.25, suggesting that the conversational interface offers more consistent usability across different search scenarios. This observation is corroborated by the rating distributions illustrated in Figure~\ref{fig:box-plot}. The latter shows that Scenario 2 received worse ratings compared to Scenario 1 for both evaluated interfaces, with the rating distributions shifting towards the lower scores, possibly indicating that Scenario 2 was slightly more difficult with regard to discovering relevant papers. This effect was especially pronounced for the graphical interface compared to the more consistent performance of the conversational interface. The more stable ratings of the conversational interface could suggest it retains usability and information relevancy, even during more challenging exploration tasks.

\paragraph{Qualitative Analysis of Participant Feedback.}
Our qualitative analysis of the free-text responses concerning the systems' strengths and weaknesses confirmed the quantitative results, revealing that the conversational interface received higher ratings and demonstrated more consistent usability than the graphical interface. Every participant was assigned an anonymous identifier between P1 and P40 to protect their privacy. In the following paragraphs, cited questionnaire responses are presented in quotation marks and assigned to their identifiers. 

The first notable strength of the conversational search interface, mentioned in nearly every second feedback comment, was the system's ease of use. For example, participant P9 remarked, ``It is easy to use and to find topics \& papers even if the prior exposure to the given topic is low.'' Users appreciated that they could immediately start talking with the conversational agent without requiring extensive knowledge of the interface or the search domain, unlike many graphical interfaces. A second strength highlighted by users was the system's guidance and structured navigation abilities, with one participant positively noting the ``Step by step process to narrow down the search and avoid search results that are not related to your query'' (P29). This feature effectively addressed the search-related vocabulary mismatches, as exemplified by the comment: ``I don't need to know exact terms or what im looking for'' (P40). Lastly, users valued the time-saving clustering and summarization features, which helped them avoid going through individual abstracts from long lists of papers. As participant P11 stated, ``[...] it understands the content of the paper and can aggregate it, without me having to manually go into the files to read the Abstract.'' These findings suggest that conversational agents can help alleviate problems associated with cognitive overload \cite{sweller1988cognitive} by gradually communicating condensed information.

Yet, a couple of participants initially struggled with understanding the three-phase search flow (e.g., ``In the first a few minutes it’s hard to understand what I can reach at the end of conversation'' (P35)). Some were also confused by the two options of selecting the suggested user response inputs displayed as buttons versus entering free-form text. This was especially the case when they wanted to reverse a choice, which participant P20 remarked, ``The options to backtracking are a bit unclear at first.'' Other participants expressed a desire to ``converse more freely'' (P9), similar to those offered by general-purpose \acp{llm}. Strengthening the integration of \acp{llm} could accommodate this preference, as \acp{llm} excel in contextual understanding of queries and navigating complex conversation logic more effectively than intent-based dialogue systems. We observed that certain users attempted to input very long requests or copy-pasting problem descriptions, an interaction more akin to \ac{llm} services like ChatGPT \cite{openai2022chat}, where users input a prompt, check the output, and refine the prompt without engaging in a proper dialogue. This type of interaction does not align with how our task-oriented dialogue system was designed to operate. Nonetheless, the evaluation shows that nearly all participants quickly figured out how our conversational system works after a few dialogue turns.

\section{Conclusion}
\label{sec:conclusion}
We proposed a conversational exploratory search system integrated with a scholarly \ac{kg}. Our study details the architectural components and presents results from training and evaluating language models that underpin the three-phase search process, including research topic classification, text clustering, and text summarization. We conducted a human evaluation to assess the system's effectiveness, identifying its perceived strengths and potential improvements. Our findings offer practical insights into the design and implementation of conversational search systems for the scholarly domain. 

\section*{Acknowledgements}
This work has been supported by the German Federal Ministry of Education and Research (BMBF) Software Campus grant 01IS17049.

\bibliography{custom}

\newpage
\onecolumn
\appendix
\section{Appendix}
\label{sec:appendix}
\begin{figure*}[h]
\centering
  \includegraphics[width=\textwidth]{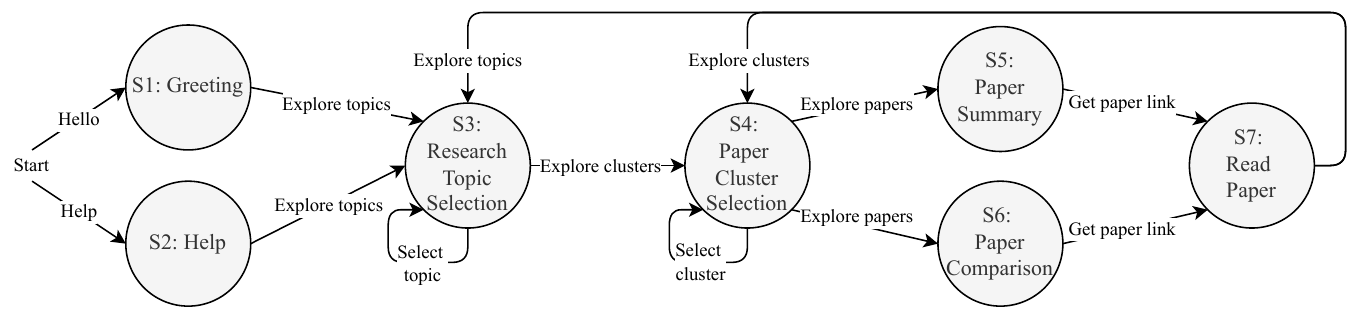}
  \caption{Conversational search flow illustrated as dialogue states (S1-S7). The three-phase search process encompasses: first, identifying a research topic (S3); second, choosing clusters of publications (S4); and third, comparing publications via short summaries (S5-S6).}
  \label{fig:dialogue-states}
\end{figure*}

\begin{figure*}[h]
  \centering
  \includegraphics[width=0.75\textwidth]{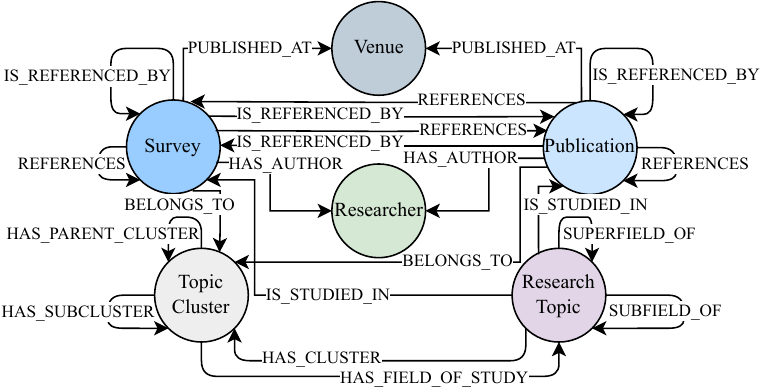}
  \caption{Semantic data model of the scholarly knowledge graph.}
  \label{fig:data-model}
\end{figure*}

The Appendix provides further insights into the results of our research, including a finite-state diagram of the conversational search flow (Figure~\ref{fig:dialogue-states}), the semantic data model of the scholarly \ac{kg} (Figure~\ref{fig:data-model}), the scenario descriptions for the human evaluation (Table~\ref{tab:scenario}), a tabular overview of the participant demographics (Table~\ref{tab:demographics}), screenshots of the conversational and graphical interface (Table~\ref{tab:screenshots}), and a collection of the crafted \ac{llm} prompts (Table~\ref{tab:prompts}).

\hfill

\hfill

\begin{table}[h]
\centering
\begin{tabular}{p{15cm}}
\hline
\textbf{Description of Scenario 1} \\
\hline
Imagine you are interested in mental health and emotional changes during the COVID pandemic. You want to analyze how people express their feelings on social media platforms during the pandemic. Your goal is to study their emotions to learn how they handle stress and anxiety. To deal with the enormous volume of data available online, you are looking for ways to automate the analysis process using NLP techniques.
\\
Your task is to use the provided search interface below to discover papers related to the described scenario. You have up to 8 minutes for your exploratory search. You are encouraged to ``think out loud''. Afterward, you will fill out an evaluation questionnaire to provide feedback on your search experience. \\
\hline
\textbf{Description of Scenario 1} \\
\hline
Imagine you are an instructor teaching a programming course in Python. You want to ensure that the exam questions reflect the learning progress of the students. Your goal is to generate multiple-choice exams according to their knowledge level. To achieve this, you are looking for ways to automatically create exam questions based on the course materials using NLP techniques.
\\
Your task is to use the provided search interface below to discover papers related to the described scenario. You have up to 8 minutes for your exploratory search. You are encouraged to ``think out loud''. Afterward, you will fill out an evaluation questionnaire to provide feedback on your search experience. \\
\hline
\end{tabular}
\caption{Scenario descriptions of the exploratory search task for the human evaluation.}
\label{tab:scenario}
\end{table}



\begin{table}[h]
\centering
\begin{tabular}{lccc}
\hline
\textbf{Demographic Variable} & \textbf{Group A (n $=$ 20)} & \textbf{Group B (n $=$ 20)} & \textbf{Overall (n $=$ 40)} \\
\hline
Mean age (age range) & 25.10 (20 to 29) & 24.95 (23 to 28) & 25.03 (20 to 29) \\
\hline
Male & 13 & 13 & 26 \\
Female & 7 & 7 & 14 \\
\hline
High school degree & 1 & - & 1 \\
Bachelor's degree & 14 & 17 & 31 \\
Master's degree & 5 & 3 & 8 \\
\hline
No \ac{nlp} knowledge & 1 & 2 & 3 \\
Beginner \ac{nlp} knowledge & 15 & 13 & 28 \\
Advanced \ac{nlp} knowledge & 4 & 5 & 9 \\
\hline
English CEFR level B1 or B2 & 3 & 5 & 8 \\
English CEFR level C1 or C2 & 17 & 15 & 32 \\
\hline
\end{tabular}
\caption{Overview of study participant demographics.}
\label{tab:demographics}
\end{table}


\begin{table}[h]
\centering
\begin{tabular}{ll}
\hline
\textbf{Conversational Interface} & \textbf{Graphical Interface} \\
\hline
\\
\includegraphics[width=0.48\textwidth, height=0.35\textwidth]{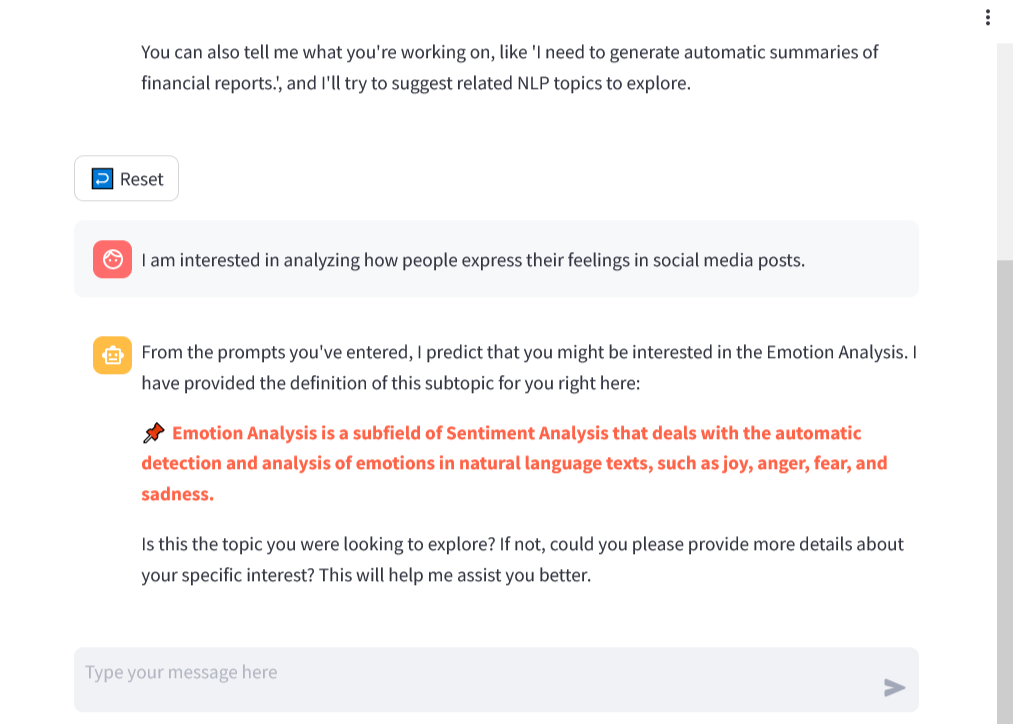} & \includegraphics[width=0.48\textwidth, height=0.35\textwidth]{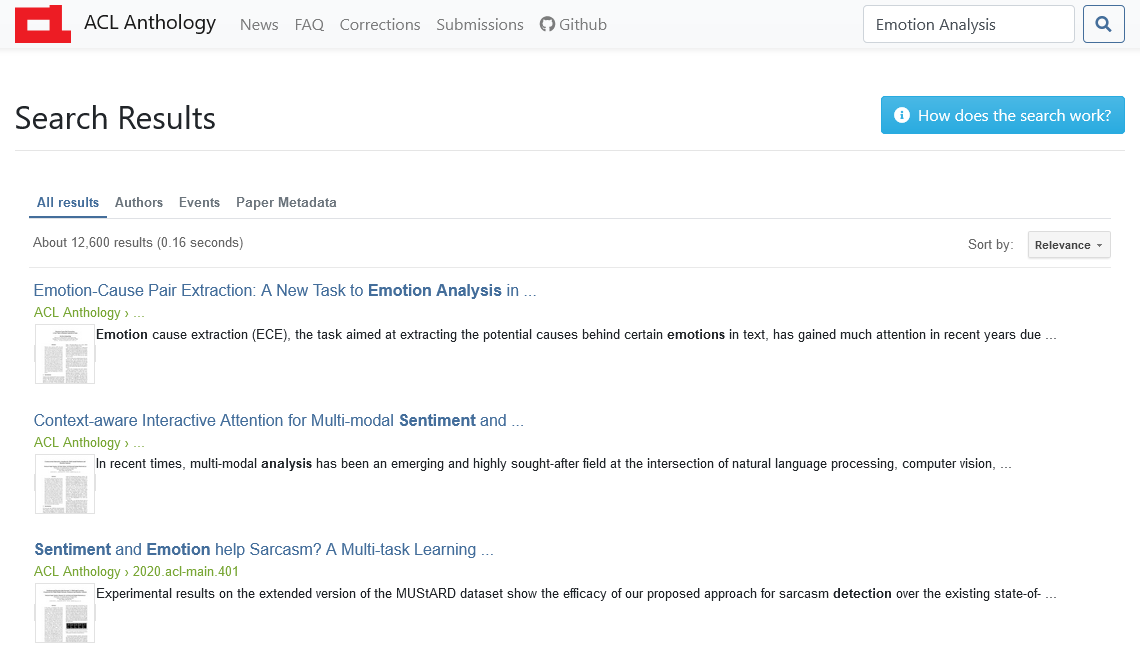} \\
\hline
\end{tabular}
\caption{Visual side-by-side comparison of the conversational and graphical interface from the human evaluation.}
\label{tab:screenshots}
\end{table}


\begin{table}[h]
\centering
\begin{tabular}{p{15cm}}
\hline
\textbf{Prompt 1: Cluster Name Generation (Zephyr-7B-Beta)} \\
\hline
Considering the themes and topics from the following TFIDF cluster tag: ``\{tfidf\_cluster\_name\}'', please provide a concise and descriptive name for a cluster that includes these \{len(paper\_list)\} academic papers: \\
\{paper\_titles\_formatted\} \\
Respond with just the cluster name, based on the overarching themes evident in the titles and the TFIDF tag. Don't include the original TFIDF cluster tag and the word ``Cluster'' in your response. \\
\hline
\textbf{Prompt 2: Comparative Text Summarization (Zephyr-7B-Beta)} \\
\hline
Please provide a comparative analysis of the objectives of two scientific papers. \\
Refer the papers with their real ids: \\
Paper \{id\_a\}'s objective is: \{obj1\} \\
Paper \{id\_b\}'s objective is: \{obj2\} \\
Highlight the key differences and similarities between Paper \{id\_a\} and Paper \{id\_b\}. \\ Use simple language. \\
Please provide a comparative analysis of the results of two scientific papers.: \\
Refer the papers with their real ids: \\
Results of Paper \{id\_a\}: \{res1\} \\
Results of Paper \{id\_b\}: \{res2\} \\
Highlight the key differences and similarities between Paper \{id\_a\} and Paper \{id\_b\}. \\ Use simple language. \\
Please provide a comparative analysis of the TLDR of two scientific papers.: \\
TLDR of Paper \{id\_a\}: \{tldr1\} \\
TLDR of Paper \{id\_b\}: \{tldr2\} \\
Highlight the key differences and similarities between Paper \{id\_a\} and Paper \{id\_b\}. \\ Use simple language. \\
\hline
\textbf{Prompt 3: LLM-Based Research Topic Classification (GPT-3.5-Turbo)} \\
\hline
You are supposed to classify a query into one of the topics provided. These topics are various fields of NLP. Your answer should be in the following format: \\
**topic name* \\
Nothing else should be included in the output. \\
Make sure there is no extra punctuation including full stops, quotation marks or anything of that sort. You are supposed to EXACTLY use the topics from the list provided. If you think it is a random question and not in the field of NLP, then return the topic as ``none''. \\
You can only provide your answer from the following topics and the topics are: \\
Multimodality \\
Natural Language Interfaces \\
Semantic Text Processing \\
Semantic Analysis \\
Syntactic Text Processing \\
Linguistic and Cognitive NLP \\
Responsible NLP \\
Reasoning \\
Multilinguality \\
Information Retrieval \\
Information Extraction and Text Mining \\
Text Generation \\
Query: \{query\}. \\
Topic:  \\
\hline
\end{tabular}
\caption{Overview of large language model prompts for various generative tasks using Zephyr-7B-Beta and GPT-3.5-Turbo. Dynamically inserted variables are enclosed within curly brackets.}
\label{tab:prompts}
\end{table}

\end{document}